\documentclass[11pt,a4paper]{article}
\usepackage[hyperref]{acl2019}
\usepackage{times}
\usepackage{latexsym}

\usepackage{url}

\aclfinalcopy 

\title{Question Answering is a Format; When is it Useful?}

\author{Matt Gardner\textsuperscript{$\spadesuit$}, Jonathan Berant\textsuperscript{$\spadesuit$,$\clubsuit$}, Hannaneh Hajishirzi\textsuperscript{$\spadesuit$,$\diamondsuit$}, \\
\textbf{Alon Talmor\textsuperscript{$\clubsuit$}, and Sewon Min\textsuperscript{$\diamondsuit$}} \\
  \textsuperscript{$\spadesuit$}Allen Institute for Artificial Intelligence \\
  \textsuperscript{$\clubsuit$}Tel Aviv University \\
  \textsuperscript{$\diamondsuit$}University of Washington \\
  \texttt{mattg@allenai.org} \\}

\begin{document}
\maketitle
\begin{abstract}
  Recent years have seen a dramatic expansion of tasks and datasets posed as question answering, from reading comprehension, semantic role labeling, and even machine translation, to image and video understanding.  With this expansion, there are many differing views on the utility and definition of ``question answering'' itself.  Some argue that its scope should be narrow, or broad, or that it is overused in datasets today.  In this opinion piece, we argue that question answering should be considered a \emph{format} which is sometimes useful for studying particular phenomena, not a phenomenon or task in itself.  We discuss when a task is correctly \emph{described} as question answering, and when a task is usefully \emph{posed} as question answering, instead of using some other format.
\end{abstract}

\section{Introduction}
``Question answering'' (QA) is a deceptively simple description of an incredibly
broad range of phenomena.  Its original use in the natural language processing
(NLP) and information retrieval (IR) literature had a very narrow scope:
answering open-domain factoid questions that a person might pose to a retrieval
system \cite{Voorhees1999TheTQ,kwok2001scaling}.  As NLP systems have improved, people have
started using question answering as a format to perform a much wider variety of
tasks, leading to a dilution of the term ``question answering''.  This dilution is natural: questions are simply a class of sentences that can have arbitrary semantics, so ``question answering'' per se also has arbitrary scope.

In this paper we aim to give some clarity to what ``question answering'' is
and when it is useful in the current NLP and computer vision literature.  Some
researchers advocate for only using question answering when the task involves
questions that humans would naturally ask in some setting
\cite{kwiatkowski2019natural,Yogatama2019LearningAE,clark2019boolq}, while others push for treating
\emph{every} NLP task, even classification and translation, as question
answering \cite{Kumar2016AskMA,McCann2018TheNL}.  Additionally, some in the NLP community have expressed fatigue at the proliferation of question answering datasets, with some conference presentations and paper submissions feeling like they need to address head-on the complaint of ``yet another QA dataset''.

We argue that ``question answering'' should be considered a \emph{format} (as opposed to other formats such as slot filling) instead of a \emph{phenomenon} or \emph{task} in itself. Question answering is mainly a useful format in tasks that require {\it question understanding} i.e.,  understanding the language of the question is a non-trivial part of the task itself (detailed in Section~\ref{sec:utility}).  If the questions can be replaced by integer ids without meaningfully altering the nature of the task, then  question understanding is not required.   Sometimes, question answering is a useful format even for datasets that do not require question understanding, and we elaborate on this in Section~\ref{sec:qa-as-transfer}.


We argue that there are three broad motivations for using question answering as a format for a particular task.  The first, and most obvious, motivation is (1) to fill human information needs, where the data is naturally formatted as question answering, because a person is asking a question.  This is not the only valid use of question answering, however.  It is also useful to pose a task as question answering (2) to probe a system's understanding of some context (such as an image, video, sentence, paragraph, or table) in a flexible way that is easy to annotate, and (3) to transfer learned parameters or model architectures from one task to another (in certain limited cases).  We give a detailed discussion of and motivation for these additional uses.

In short: question answering is a format, not a task.  Calling a dataset a ``QA
dataset'' is not an informative description of the nature of the task, nor is it meaningful to talk about ``question answering datasets'' as a cohesive group without some additional qualifier describing what \emph{kind} of question answering is being performed.  The community should not be concerned that many different datasets choose to use this format, but we \emph{should} be sure that all datasets purporting to be ``QA datasets'' are in fact reasonably described this way, and that the choice of question answering as a format makes sense for the task at hand.

\section{Question Answering is a Format}
\label{sec:definition}

A question is just a particular kind of natural language sentence.  The space of
things that can be asked in a natural language question is arbitrarily large.
Consider a few examples: \textit{``What is 2 + 3?"},  \textit{``What is the
sentiment of this sentence?"},  \textit{``What's the fastest route to the
hospital?"}, and \textit{``Who killed JFK?"}. It seems ill-advised to treat all of these questions as conceptually similar; just about the only unifying element between them is the question mark.  ``Question answering'' is clearly not a cohesive phenomenon that should be considered as a ``task'' by NLP researchers.

What, then, is question answering?  It is a \emph{format}, a way of posing a particular problem to a machine, just as classification or natural language inference are formats.  The phrase ``yet another question answering dataset'' is similar in meaning to the phrase ``yet another classification dataset''---both question answering and classification are \emph{formats} for studying particular phenomena.  Just as classification tasks differ wildly in their complexity, domain, and scope, so do question answering tasks.  Question answering tasks \emph{additionally} differ in their output types (consider the very different ways that one would provide an answer to the example questions above), so it is not even a particularly unified format.  The community should stop thinking of ``question answering'' as a \emph{task} and recognize it as a \emph{format} that is useful in some situations and not in others. Instead, the community should consider finding useful cases of whether to pose a task as a question answering format or not. 

Question answering is mainly useful in tasks where understanding the question language is itself part of the task.  This typically means that the phenomena being queried (i.e., the questions in the dataset) do not lend themselves well to enumeration, because the task is either unbounded or inherently compositional.  If every question in the data can be replaced by an integer id without fundamentally changing the nature of the task, then it is usually not useful to pose the task as question answering. 

To demonstrate this point, we begin with an extreme example.  Some works treat
traditional classification or tagging problems as question answering, using
questions such as \emph{``What's the sentiment?''} or \emph{``What are the POS
tags?''}
\citep{Kumar2016AskMA,McCann2018TheNL}.  In these cases, not only can every
question be replaced by a single integer, they can all be replaced by the
\emph{same} integer.  There is no meaningful question understanding component in
this formulation.  This kind of reformulation of classification or tagging as question
answering is occasionally useful, but only in rare circumstances when trying to
transfer models across related datasets (Section \ref{sec:qa-as-transfer}). 

As a less extreme example, consider the WikiReading dataset
\citep{Hewlett2016WikiReadingAN}.  In this dataset, a system must read a
Wikipedia page and predict values from a structured knowledge base.  The type of
the value, or ``slot'' in the knowledge base, can be represented by an integer
id.  One could also pose this task as question answering, by writing a question
template for each slot.  These templates, however, are easily memorized by a
learning system given enough data, meaning that understanding the language in the templates is not a significant part of the underlying task.  The template could be replaced by the integer id of the slot without changing the task; the purpose of the template is largely for humans to understand the example, not the machines.\footnote{Or for transfer from a QA dataset; c.f. QA-ZRE \citep{Levy2017ZeroShotRE}, discussed in \S\ref{sec:qa-as-transfer}.}  Attempts to have multiple surface realizations of each template do not help here; the system can still memorize template cluster ids.  Even when formatted as question answering, we argue this kind of dataset is more appropriately called ``slot filling''.  Similarly, a dataset with a question template that involves an entity variable (e.g., \textit{When was [PERSON] born?}) is simply a pairing of an integer id with an entity id, and does not require meaningful question understanding.  This is still appropriately called ``slot filling''. 

Some templated language datasets can be considered ``question answering'', however.  The CLEVR and GQA datasets \cite{Johnson2017CLEVRAD,Hudson2019GQAAN}, for example, use synthetic questions generated from a complex grammar.  While these certainly aren't \emph{natural language} questions, the dataset is still requires question understanding, because the questions are complex, compositional queries and replacing them with single integer ids misses the opportunity of modeling the compositional structure and dramatically reducing sample complexity. There is admittedly a fuzzy line between complex slot filling cases with multiple variables and grammar-based templated generation, but we believe the basic principle is still valid: if it is reasonable to solve the problem by assigning each question an id (or an id paired with some variable instantiations), then the task does not require significant question understanding and is likely more usefully posed with a format other than question answering.

\section{When QA is useful}
\label{sec:utility}

In the previous section we argued that question answering is best thought of as a \emph{format} for posing particular tasks, and we gave a concrete definition for when a task should be \emph{called} question answering.  In this section we move to a discussion of when this format is \emph{useful}.

There are three very different motivations from which researchers arrive at
question answering as a format for a particular task, all of which are
potentially useful.  The first is when the goal is to fill human information
needs, where the end task involves humans naturally asking questions.  The
second is when the complexity or scope of a task go beyond the bounds of a fixed
formalism, requiring the flexibility of natural language as an annotation and/or
querying mechanism.  The third is that question answering, or a unified input/output format in general, might be a way to transfer learned knowledge from one task to a related task.

\subsection{Filling human information needs}

There are many scenarios where people naturally pose questions to machines,
wanting to receive an answer.  These scenarios are too varied to enumerate, but
a few examples are search
queries~\citep{dunn2017searchqa,kwiatkowski2019natural}, natural language
interfaces to databases
\cite{zelle96geoquery,berant2013semantic,iyer2017neural}, and virtual assistants
\cite{dahl1994expanding}. In addition to practical usefulness, natural questions
prevent some of the biases found in artificial settings, as analyzed by \newcite{lee2019latent} (though they will naturally have their own biased distribution). These are ``natural'' question answering settings, and keeping the natural format of the data is an obvious choice that does not need further justification, so we will not dwell on this section.  The danger is to think that this is the \emph{only} valid use of question answering as a format.  It is not, as the next two sections will show.

\subsection{QA as annotation / probe}

When considering building a dataset to study a particular phenomenon, a researcher has many options for how that dataset should be formatted.  The most common approach in NLP is to define a formalism, typically drawn from linguistics, and train people to annotate data in that formalism (e.g., part of speech tagging, syntactic parsing, coreference resolution, etc.).  In computer vision research, a similar approach is taken for image classification, object detection, scene recognition, etc.  When the phenomenon being studied can be cleanly captured by a fixed formalism, this is a natural approach.

There are times, however, when defining a formalism is not optimal.  This can be either because the formalism is too expensive to annotate, or because the phenomenon being annotated does not fit nicely into a fixed formalism.  In these cases, the flexibility and simplicity of \emph{natural language} annotations can be incredibly useful.  For example, researchers often rely on crowd workers when constructing datasets, and training them in a linguistic formalism can be challenging.  However, there are many areas of semantics that any native speaker could easily annotate in natural language, without needing to be taught a formalism (c.f. QA-SRL, described below).

Having decided on natural language annotations instead of a fixed formalism still leaves a lot of room for choice of formats.  Free-form generation, such as in image captioning and summarization, or natural language inference, are also flexible formats that use natural language as a primary annotation mechanism \cite{Poliak2018CollectingDN}.  In what circumstances should one use question answering instead of these other options?

Question answering is often a good choice over summarization or captioning-style formats when (1) there are many things about a given context that could be queried.  In summarization and captioning, only one output per input image or passage is generated.  Question answering allows the dataset designer to query several different aspects of the context.  Question answering may also be preferred over summarization-style formats (2) for easier evaluation.  Current metrics for automatically evaluating natural language generation are not very robust \cite[\emph{inter alia}]{Edunov2019OnTE}.  In question answering formats, restricted answer types, such as span extraction, are often available with more straightforward evaluation metrics, though those restrictions often come with their own problems, such as reasoning shortcuts or other biases \cite{jia-liang-2017-adversarial,min-etal-2019-compositional}.

Question answering is strictly more general than natural language inference (NLI) as a format, as an NLI example can always be converted to a QA example by phrasing the hypothesis as a question and using yes, no, or maybe as the answer.  The opposite is not true, as questions with answers other than yes, no, or maybe are challenging to convert to NLI format without losing information.  The question and answer can be converted into a declarative hypothesis with label ``entailed'', but coming up with a useful negative training signal is non-trivial and very prone to introducing biases.  Because the output space is larger in QA, there is a richer learning signal from each example.  We recommend using QA over NLI as a format for new datasets in almost all cases.

The remainder of this section looks at specific examples (in a non-exhaustive
manner) where question answering is usefully used as an annotation mechanism for particular phenomena.  In none of these cases would a human seeking information actually ask any of the questions in the dataset; the person would just look at the given context (sentence, image, paragraph, etc.) to answer their own question.  There is no "natural distribution" of questions for this kind of task.

\paragraph{QASRL / QAMR}

The motivation for this work is explicitly to make annotation cheaper by having crowd workers label semantic dependencies in sentences using natural language instead of training them in a formalism~\citep{he2015question,fitzgerald2018large,michael-etal-2018-crowdsourcing}. In addition, the QA pairs can also be seen as a probe for understanding the structure of the sentence.

\paragraph{Visual question answering}

The motivation for this task is to evaluate understanding images and
videos---the point is that QA allows capturing a much richer set of phenomena
than using a fixed formalism~\citep{antol2015vqa,Zellers_2019_CVPR}.

\paragraph{Reading comprehension}
The motivation for this task is to  demonstrate understanding of a passage of text, using various kinds of questions~\citep{Rajpurkar2016SQuAD10,Joshi2017TriviaQAAL,dua2019drop,amini2019mathqa}. These questions aim at evaluating different phenomena, from understanding simple relationships about entities, to numerical analysis, to multi-hop reasoning.  There are two recent surveys that give a good overview of the kinds of reading comprehensions that have been built so far \cite{Liu2019NeuralMR,Zhang2019MachineRC}.  The open-domain nature of the reading comprehension task makes it very unlikely that a formalism could be developed, leaving question answering as the natural way to probe a system's understanding of longer passages of text.

\paragraph{Background knowledge and common sense}
The motivation for this task is to probe whether a machine exhibits the type of
background knowledge and common sense exhibited by a typical human
\cite{mihaylov2018openbookqa,talmor2019commonsenseqa,sap2019socialiqa}, which, again, is difficult to
capture with a fixed formalism.

\subsection{As a transfer mechanism}
\label{sec:qa-as-transfer}

There has been a lot of work on transferable representation learning, trying to share linguistic or other information learned between a diverse set tasks.  The dominant, and most successful, means of doing this is by sharing a common language representation layer, and having several different task-specific heads that output predictions in particular formats.  An alternative approach is to pose a large number of disparate tasks in the same format.  This has generally been less successful, though there are a few specific scenarios in which it appears promising. Below, we highlight two of them.

The first case in which it helps to pose a non-QA task as QA is when the non-QA task is closely related to a QA task, and one can reasonably hope to get few-shot (or even zero-shot) transfer from a trained QA model to the non-QA task.  This \emph{model transfer} was successfully demonstrated by \citet{Levy2017ZeroShotRE}, who took a relation extraction task and used templates to pose it as QA in order to transfer a SQuAD-trained model to the task.  However, as in other transfer learning or multi-task learning scenarios, this is only likely to succeed when the source and target tasks are very similar to each other.  \citet{McCann2018TheNL} attempted to do multi-task learning with many diverse tasks, including machine translation, summarization, sentiment analysis, and more, all posed as question answering.  In most cases, this hurt performance over training on each task independently.  It seems likely that having a shared representation layer and separate prediction heads would be a more fruitful avenue for achieving this kind of transfer than posing everything as QA.

The second case in which it helps to pose a non-QA task as QA is when the model \emph{architectures} used in QA are helpful for the task.
\citet{das2018building} and \citet{li2019entity} achieve significant improvement
by converting the initial format of their data (entity tracking and relation extraction,
respectively) to a QA format via question templates and using a QA model with no
pretraining. We hypothesize that in these cases, forcing the model to compute similarities between the input text and the words in the human-written templates provides a useful inductive bias. For example, the template \emph{``Where
is CO2 created?"} will encourage the model to map \emph{``where''} to locations in the passage and to find
synonyms of \emph{``CO2''}, inductive biases which may be difficult to inject in other model architectures.

\section{Conclusion}

In this paper, we argued that the community should think of question answering as a format, not a task, and that we should not be concerned when many datasets choose to use this format.  Question answering is a useful format predominantly when the task has a non-trivial \emph{question understanding} component, and the questions cannot simply be replaced with integer ids.  We observed three different situations in which posing a task as question answering is useful: (1) when filling human information needs, and the data is already naturally formatted as QA; (2) when the flexibility inherent in natural language annotations is desired, either because the task does not fit into a formalism, or training people in the formalism is too expensive; and (3) to transfer learned representations or model architectures from a QA task to another task.  As NLP moves beyond sentence-level linguistic annotation, many new datasets are being constructed, often without well-defined formalisms backing them.  We encourage those constructing these datasets to think carefully about what format is most useful for them, and we have given some guidance about when question answering might be appropriate.

\bibliography{acl2019}
\bibliographystyle{acl_natbib}

\end{document}